\documentclass[journal]{vgtc}                
\ifpdf
  \pdfoutput=1\relax                   
  \usepackage[]{graphicx}                
  \DeclareGraphicsExtensions{.pdf,.png,.jpg,.jpeg} 
\else
  \usepackage{graphicx}                
  \DeclareGraphicsExtensions{.eps}     
\fi%

\onlineid{0}

\vgtccategory{Research}
\vgtcpapertype{algorithm/technique}

\title{Deep Volumetric Ambient Occlusion}

\author{Dominik Engel and Timo Ropinski}
\authorfooter{
\item
 Dominik Engel is with Ulm University. E-Mail: dominik.engel@uni-ulm.de
\item
 Timo Ropinski is with Ulm University and Linköping University (Norrköping). E-Mail: timo.ropinski@uni-ulm.de
}

\shortauthortitle{Engel \MakeLowercase{\textit{et al.}}: Deep Volumetric Ambient Occlusion}

 \abstract{We present a novel deep learning based technique for volumetric ambient occlusion in the context of direct volume rendering. Our proposed \emph{Deep Volumetric Ambient Occlusion} (DVAO) approach can predict per-voxel ambient occlusion in volumetric data sets, while considering global information provided through the transfer function. The proposed neural network only needs to be executed upon change of this global information, and thus supports real-time volume interaction. Accordingly, we demonstrate DVAO's ability to predict volumetric ambient occlusion, such that it can be applied interactively within direct volume rendering. To achieve the best possible results, we propose and analyze a variety of transfer function representations and injection strategies for deep neural networks. Based on the obtained results we also give recommendations applicable in similar volume learning scenarios. Lastly, we show that DVAO generalizes to a variety of modalities, despite being trained on computed tomography data only.}


\CCScatlist{ 
 \CCScat{K.6.1}{Management of Computing and Information Systems}%
{Project and People Management}{Life Cycle};
 \CCScat{K.7.m}{The Computing Profession}{Miscellaneous}{Ethics}
}

\usepackage{tikz}
\begin{document}
\begin{tikzpicture}[remember picture,overlay]
    \node[inner sep=0pt] at (current page.center) {\includegraphics[page=1]{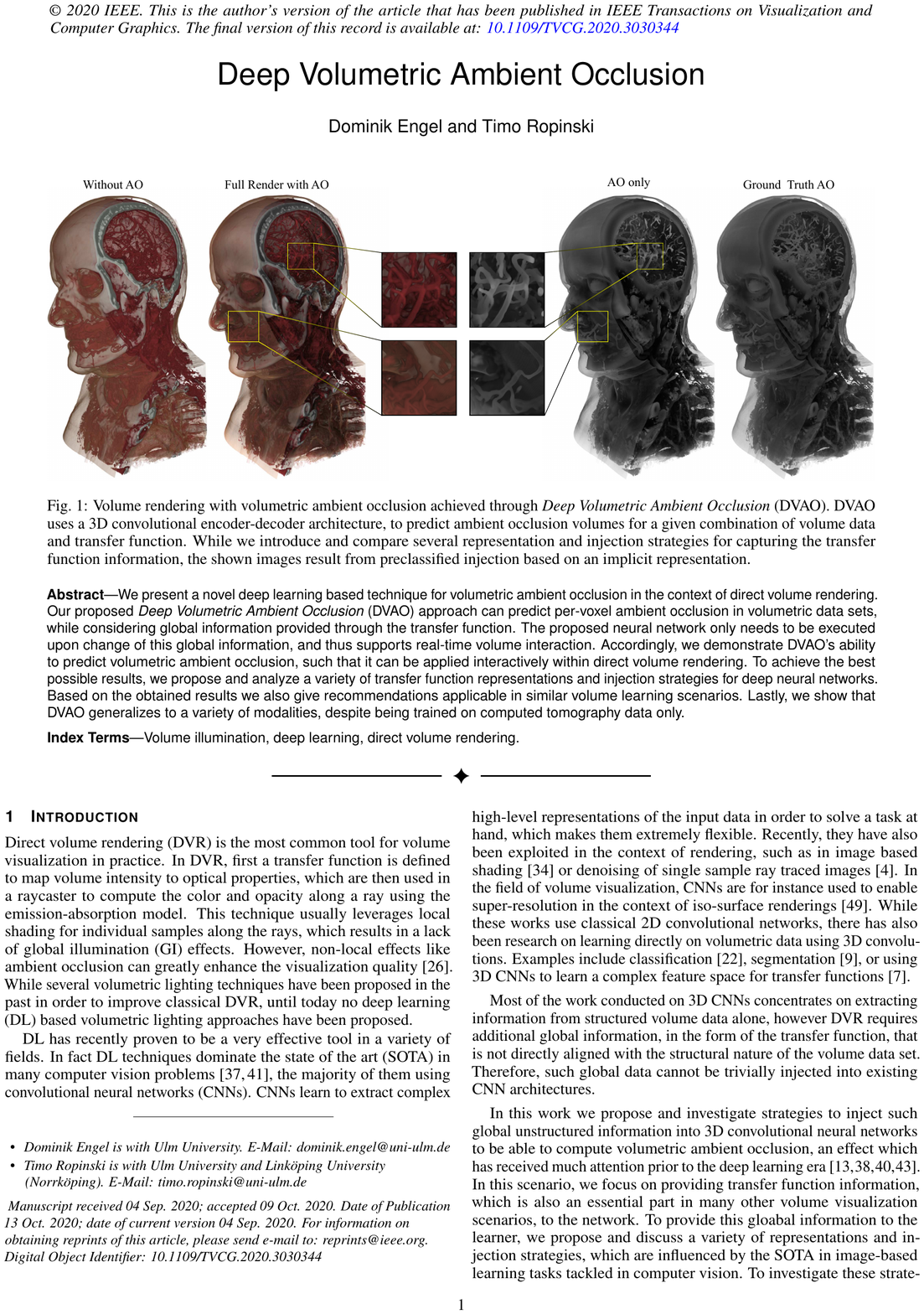}};
\end{tikzpicture}
\newpage
\clearpage
\begin{tikzpicture}[remember picture,overlay]
    \node[inner sep=0pt] at (current page.center) {\includegraphics[page=2]{dvao}};
\end{tikzpicture}
\newpage
\clearpage
\begin{tikzpicture}[remember picture,overlay]
    \node[inner sep=0pt] at (current page.center) {\includegraphics[page=3]{dvao}};
\end{tikzpicture}
\newpage
\clearpage
\begin{tikzpicture}[remember picture,overlay]
    \node[inner sep=0pt] at (current page.center) {\includegraphics[page=4]{dvao}};
\end{tikzpicture}
\newpage
\clearpage
\begin{tikzpicture}[remember picture,overlay]
    \node[inner sep=0pt] at (current page.center) {\includegraphics[page=5]{dvao}};
\end{tikzpicture}
\newpage
\clearpage
\begin{tikzpicture}[remember picture,overlay]
    \node[inner sep=0pt] at (current page.center) {\includegraphics[page=6]{dvao}};
\end{tikzpicture}
\newpage
\clearpage
\begin{tikzpicture}[remember picture,overlay]
    \node[inner sep=0pt] at (current page.center) {\includegraphics[page=7]{dvao}};
\end{tikzpicture}
\newpage
\clearpage
\begin{tikzpicture}[remember picture,overlay]
    \node[inner sep=0pt] at (current page.center) {\includegraphics[page=8]{dvao}};
\end{tikzpicture}
\newpage
\clearpage
\begin{tikzpicture}[remember picture,overlay]
    \node[inner sep=0pt] at (current page.center) {\includegraphics[page=9]{dvao}};
\end{tikzpicture}
\newpage
\clearpage
\begin{tikzpicture}[remember picture,overlay]
    \node[inner sep=0pt] at (current page.center) {\includegraphics[page=10]{dvao}};
\end{tikzpicture}
\newpage
\clearpage
\begin{tikzpicture}[remember picture,overlay]
    \node[inner sep=0pt] at (current page.center) {\includegraphics[page=11]{dvao}};
\end{tikzpicture}
\newpage
\clearpage
\begin{tikzpicture}[remember picture,overlay]
    \node[inner sep=0pt] at (current page.center) {\includegraphics[page=1]{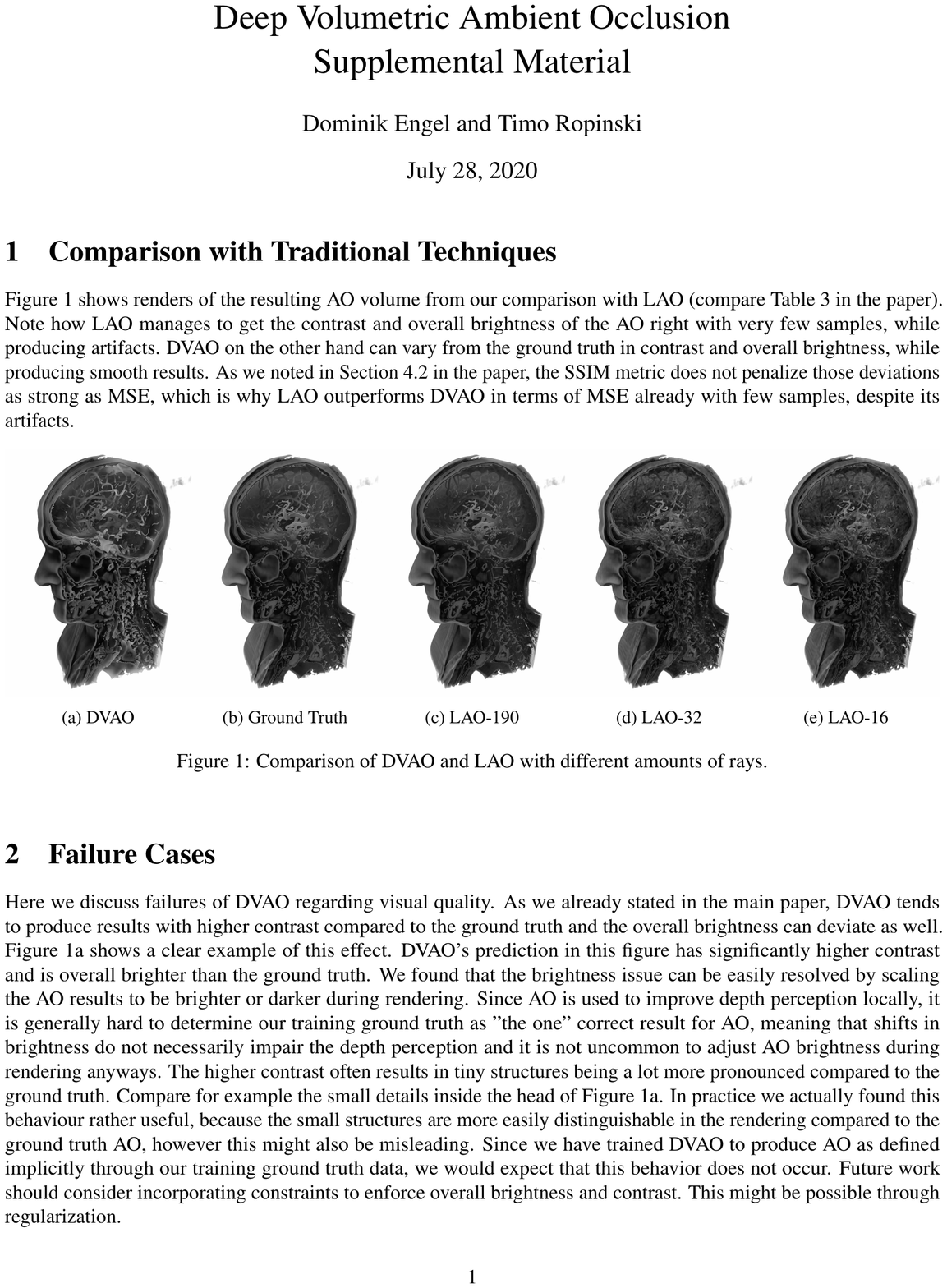}};
\end{tikzpicture}
\newpage
\clearpage
\begin{tikzpicture}[remember picture,overlay]
    \node[inner sep=0pt] at (current page.center) {\includegraphics[page=2]{supmat}};
\end{tikzpicture}
\newpage
\clearpage
\begin{tikzpicture}[remember picture,overlay]
    \node[inner sep=0pt] at (current page.center) {\includegraphics[page=3]{supmat}};
\end{tikzpicture}
\newpage
\clearpage
\begin{tikzpicture}[remember picture,overlay]
    \node[inner sep=0pt] at (current page.center) {\includegraphics[page=4]{supmat}};
\end{tikzpicture}
\end{document}